\documentclass[letterpaper, 10 pt, journal, twoside]{IEEEtran} 

\usepackage{amsmath} 
\usepackage{amssymb}  
\usepackage{bm}

\usepackage{hyperref}
\hypersetup{
    colorlinks = true,
    citecolor = blue,
	urlcolor = black,
}

\usepackage[pdftex]{graphicx}

\begin{document}
\title{Thermal Recovery of Multi-Limbed Robots with Electric Actuators}

\author{Steven Jens Jorgensen$^{1, 2}$, James Holley$^{2}$, Frank Mathis$^{2}$, \\ Joshua S. Mehling$^{2}$, and Luis Sentis$^{1}$
\thanks{Manuscript received: September, 10, 2018; Revised November 28, 2018; Accepted December, 12, 2018.}
\thanks{This paper was recommended for publication by Editor Nikos Tsagarakis upon evaluation of the Associate Editor and Reviewers' comments. This work was supported by a NASA Space Technology Research Fellowship (NSTRF) Grant \# NNX15AQ42H, and the Office of Naval Research, ONR grant \#N000141512507.} 
\thanks{$^{1}$ The authors are with the University of Texas at Austin.}%
\thanks{$^{2}$ The authors are with the NASA Johnson Space Center.}%
\thanks{Digital Object Identifier (DOI): 10.1109/LRA.2019.2894068}
}

\markboth{IEEE Robotics and Automation Letters. Preprint Version. Accepted December, 2018. Arxiv Corrections May, 2019}
{Jorgensen \MakeLowercase{\textit{et al.}}: Thermal Recovery of Multi-Limbed Robots}  

\maketitle

\begin{abstract}
The problem of finding thermally minimizing configurations of a humanoid robot to recover its actuators from unsafe thermal states is addressed. A first-order, data-driven, effort-based, thermal model of the robot's actuators is devised, which is used to predict future thermal states. Given this predictive capability, a map between configurations and future temperatures is formulated to find what configurations, subject to valid contact constraints, can be taken now to minimize future thermal states. Effectively, this approach is a realization of a contact-constrained thermal inverse-kinematics (IK) process. Experimental validation of the proposed approach is performed on the NASA Valkyrie robot hardware.
\end{abstract}


\section{Introduction}
\IEEEPARstart{E}{ffective} thermal management is necessary for continuous long-term deployment of robots in ground \cite{han2016protection} and space applications \cite{swanson2003nasa}. At NASA Johnson Space Center (JSC), the operation of the Valkyrie humanoid robot \cite{radford2015valkyrie} constantly requires the operator to monitor the thermal states of the robot's electric actuators. When the temperatures of the torso or leg actuators reach unsafe levels, robot operations are postponed until the thermal states of the actuators return within safe limits. Existing techniques to manage the thermal state of a robot fall into three categories: considering heat dissipation in the design \cite{seok2015design} \cite{kenneally2016design}, the use of novel materials for insulation against external heat \cite{han2016protection}, and active methods with liquid cooling \cite{swanson2003nasa, urata2008thermal, paine2015design}. However, these approaches are only appropriate during the design and manufacturing phase of the robot as modifying the hardware of existing robots, such as Valkyrie, is difficult.

\begin{figure}
\centering
\includegraphics[width=\columnwidth]{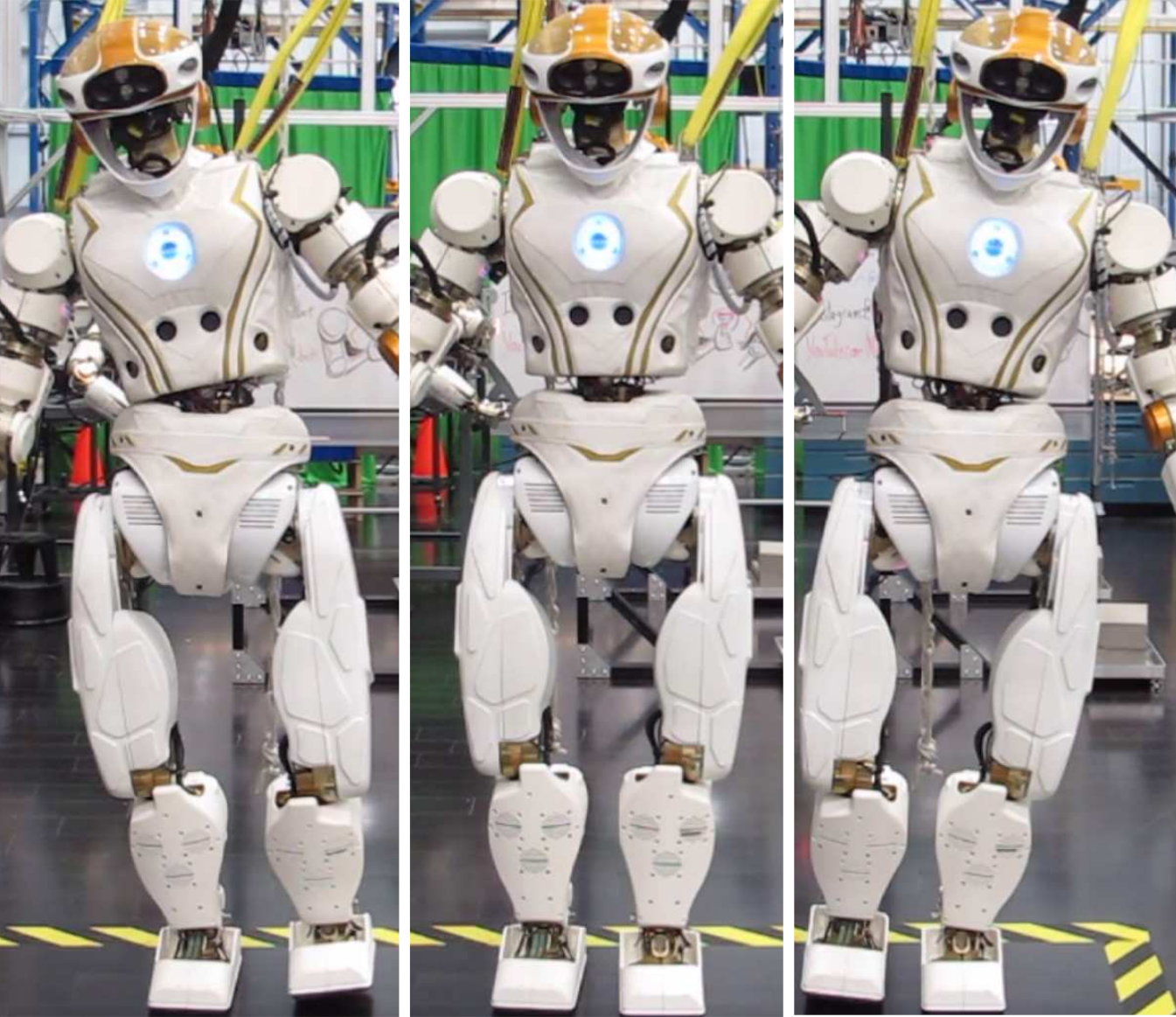}
\caption{Thermally minimizing configurations for three different contact configurations found by the proposed algorithm for actuator thermal recovery. From left to right, the valid contact configurations explored in this paper are: right leg single support, double support, and left leg single support. During the thermal recovery process, the algorithm naturally finds a strategy that switches between these thermally minimizing configurations.}
\label{fig:thermally_minimizing_poses}
\end{figure}

One thermal recovery strategy for certain types of robots is to shut off all the motors. This strategy is appropriate for robots with high gear ratios such that the majority of the robot load is supported by the mechanical structure of the system. However, this approach can be impractical for multi-limbed robots such as humanoids, which need to exert effort to balance \cite{radford2015valkyrie, englsberger2014overview, dafarra2016torque}. For these robots, a zero effort condition is similar to being in a fallen state, and standing up from a fallen state remains a hard problem \cite{krotkov2017darpa, guizzo2015hard}. The robot may also be constrained to a particular set of valid contact configurations, and in these cases the robot is required to exert motor effort to satisfy contact constraints.

In contrast to redesigning the robot or augmenting the hardware with active liquid cooling components, the described approach searches for thermally minimizing robot configurations under different valid contact constraints to thermally recover the actuator states. Concretely, a data-driven, effort-based (forces and torques), thermal model of Valkyrie's actuators is first created. This thermal model is used to predict future temperatures given actuator efforts. Then, a mapping between contact-consistent configurations and actuator efforts is formulated using constrained dynamics equations \cite{sentis2007synthesis}. This enables the construction of a potential function on the thermal state vector in terms of robot configuration. Finally, a contact-consistent gradient descent is performed on the potential function which finds a thermally minimizing configuration for a given contact constraint (Fig.~\ref{fig:thermally_minimizing_poses}). This minimization process is a realization of thermal-based inverse-kinematics (IK). An experimental hardware validation is performed on the Valkyrie robot, which demonstrates that the presented algorithm is able to recover the actuators from unsafe thermal states. It is also shown that a thermal-aware contact-switching strategy can recover dangerously warm actuators faster than employing a minimum-effort strategy only.

The contributions of this work are the following: (1) A detailed description of the effort-based thermal model and its system identification process. (2) The formulation of a temperature potential function in terms of configuration under contact constraints. (3) A gradient-descent approach to find contact-consistent configurations which minimizes the temperature potential function. 

\section{Related Works}
\subsection{Actuator Thermal Modeling}
For thermal system identification, the work presented in \cite{paine2015design}, which is inspired from \cite{urata2008thermal}, used second order dynamics to model the thermal parameters of the motor core and the motor case. Utilizing the reported thermal parameters of the motor core from the manufacturer's specification sheet, a step-response test was used to obtain the thermal resistance and capacitance of the motor case. In this previous work, it was important to estimate the internal temperature of the motor core. However, Valkyrie's thermistor configuration (see Sec.~\ref{sec:val_robot_system}) already provides the desired thermal states, and because the proposed approach uses a thermal model to predict the evolution of the thermal states, it is possible to use a simplified first-order model similar to \cite{stemme1994principles}. 

While the system identification process of \cite{paine2015design} was performed on a table-top platform, in contrast, the system identification described here is performed while the actuators are still in the robot. Due to the difficulty of getting good step responses for each actuator in the robot, the thermal system identification method instead utilizes a least-squares fit via gradient descent. This has the advantage of learning the thermal parameters from large amounts of data with the added benefit that parameter identification will not be restricted to step response properties. A gradient descent approach also enables adaptive online parameter identification (eg: with stochastic gradient descent) \cite{duchi2011adaptive}, as mini-batches of new data can be easily incorporated to update environmentally sensitive thermal parameters.

\subsection{Deriving Configurations from Minimum Effort Control}
Since thermal models of electric actuators are directly driven by the output effort of the actuator \cite{stemme1994principles}, one simple approach for thermal recovery of actuators is to simply identify minimum effort configurations from torque commands. In \cite{sentis2007synthesis}, a task Jacobian which described the relationship of gravity to joint configuration was incorporated to a whole-body operational space controller, which produced minimum effort torques while satisfying balance constraints. Similarly, a quadratic-program (QP) formulation such as in \cite{aghili2016control} can provide minimizing torques while satisfying contact constraints. However, since these approaches do not have thermal information on the actuators, a minimizing configuration from a torque output will not necessarily recover thermal states, as the minimizing configuration itself can still heat actuators with unsafe thermal states. Suppose a quadruped robot has just completed a complex locomotion task which required large amounts of effort from a single leg and no further load must be put on this limb. A minimum effort only strategy will load this actuator during thermal recovery. To address this problem, this work extends \cite{sentis2007synthesis} by first directly minimizing thermal states instead of gravity effort, second by obtaining configurations during contact-consistent gradient-descent instead of joint torque outputs from an operational space mapping, and third by searching over thermally minimizing configurations from a set of valid contact configurations. 

\section{The Valkyrie Robot System}
\label{sec:val_robot_system}
The NASA Valkyrie robot \cite{radford2015valkyrie} is used for experimental validation. Excluding the finger joints of the robot, Valkyrie has 32 actuated degrees of freedom. The robot is also comprised of harmonic-driven, series-elastic electric actuators \cite{pratt1995series,paine2014design}, for torque output sensing and control. Each electric actuator contains three thermistors for monitoring the temperature of the actuator controller's logic board, the temperature of the motor driver's/bridge's heat sink, and the temperature of the motor core. The measured temperatures from the thermistors are recorded and broadcast on the operator's console for 
thermal state monitoring. Valkyrie's primary thermal concerns are the thermal states of the torso and leg actuators' motor drivers and cores. These components naturally heat up due to the high actuator efforts needed to balance, thus the thermal recovery work presented here focuses on these actuators.

While most of Valkyrie's joints are rotary, the torso, wrist, and ankle joints each have two linear actuators that control the joint's roll and pitch degrees of freedom. As the thermal model is based on actuator effort (see Sec.~\ref{sec:thermal_model_and_sys_id}), it is necessary to derive actuator efforts from joints with kinematic loops. For a given roll, $\Psi$, and pitch, $\theta$, configurations of the torso, wrist, and ankle joints, an analytical Jacobian describes the mapping between the output roll and pitch torques, $\bm{\tau}_{\rm{rotary}}$, and the linear forces, $\mathbf{F}_{\rm{linear}}$, of the push rods.

\begin{figure}
\centering
\includegraphics[width=\columnwidth]{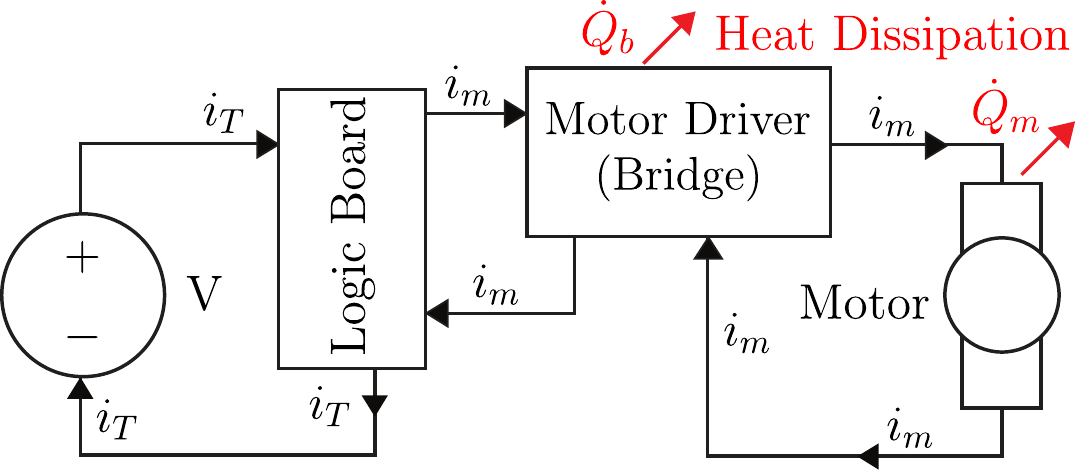}
\caption{A simplified electrical model of Valkyrie's actuator is shown depicting the relationship between the motor driver, the electric motor, and their respective heat dissipation, $\dot{Q}_b$, and $\dot{Q}_m$. A logic board with a low-level torque controller draws a total current, $i_T$, from a DC voltage source, $V$. Given a desired torque, the logic board provides the motor driver a current, $i_m$. While the motor driver and electric motor are two separate thermal systems, the same current $i_m$ passes through the driver and the motor.}
\label{fig:simplified_electrical_model}
\end{figure}

\section{Thermal Modeling and System Identification of Valkyrie's Actuators}
\label{sec:thermal_model_and_sys_id}
\subsection{Effort-based Thermal Model}
Each actuator of Valkyrie has two thermal systems of interest: the actuator's motor driver and the actuator's motor core itself. It is  assumed that the thermistors are representative of the true thermal states of the motor driver and core. Thus, what is unknown is the heat dissipation properties of each thermal system. 

\begin{figure}
\includegraphics[width=\columnwidth]{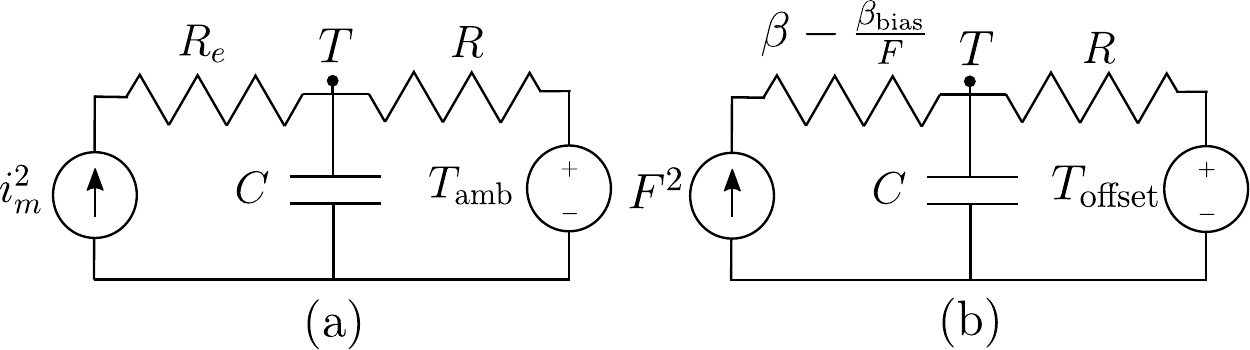}
\caption{Two different first-order thermal models for the motor driver and motor core systems. Subfigure(a) is a standard first-order thermal model for electrical systems in which the Joule heating due to the electrical current is used as the input to the thermal system. On the otherhand, subfigure(b) models the actuator effort in torque or force units as the input to the thermal system with a variable thermal resistance to model actuator bias.} 
\label{fig:current_vs_effort_thermal_models}
\end{figure}

While the approach relies on modeling the thermal parameters of the system, many of the thermal parameters are inherently sensitive to the environment which makes exact parameter identification difficult: electrical winding resistance can vary with temperature and thermal resistance depends on ventilation. Additionally, since the motor driver is a small electrical board, it can be located anywhere on the robot. Depending on its mounting configuration, it may have different thermal dissipation properties compared to a tabletop setting. Similarly, each actuator may also have different mounting and enclosure configurations, which can also lead to different thermal parameters. To address parameter sensitivity concerns, the thermal system is modeled from large amounts of actual operation data with different excitation modes and uses a first-order lumped model to enable the dominating parameters and environmental factors to dictate the thermal evolution of the model. Also note that this thermal model only needs to be able to predict the general trend of thermal evolution and relative thermal magnitudes for the thermal recovery algorithm to make informed decisions, so it does not need to be exact.

Fig.~\ref{fig:simplified_electrical_model} illustrates that for a given motor torque, the same current passes through the motor driver and core. Next, Fig.\ref{fig:current_vs_effort_thermal_models}(a) demonstrates this first-order model in which the Joule heating due to the electrical current is the input to the thermal system with thermal resistance $R$ and capacitance $C$. This has the corresponding differential equation,
\begin{align}
RC \ \dot{T}(t) + T(t) = i_m^2 R_e R + T_{\text{amb}},
\end{align}
where $i_m$ is the electrical current going through the motor driver and motor core, $R_e$ is the electrical resistance of the motor core, and $T_\text{amb}$ is the ambient temperature. 

\begin{figure}
\includegraphics[width=\columnwidth]{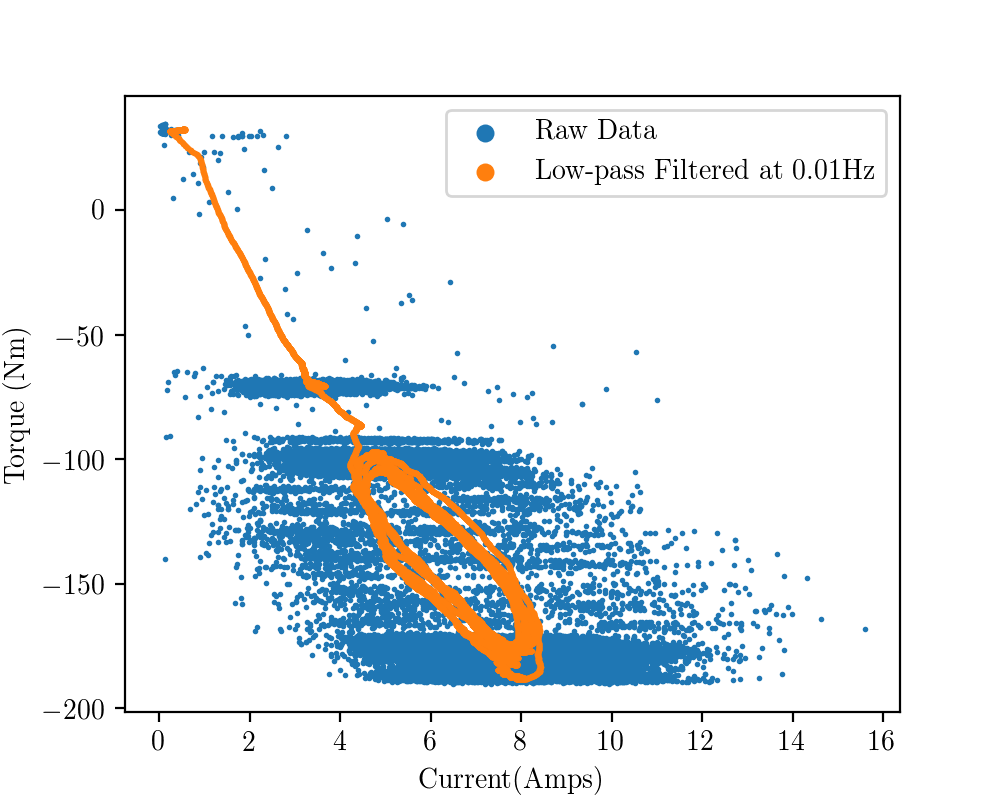}
\caption{The torque-current relationship of the harmonic-driven, series-elastic-actuated knee joint of Valkyrie. Valkyrie's arm, torso, and leg actuators have bias and hysteresis which affect the relationship between the current input to the motor and the force/torque output of the actuator. The bias arises from the spring load, and hysteresis depends on the the loading direction and actuator effort direction.} 
\label{fig:sea_bias_and_hysteresis}
\end{figure}

One difficulty working with Valkyrie's actuators with harmonic drives and series-elastic components is that its compliance introduces a bias and hysteresis on the torque-current relationship of the actuator \cite{ruderman2012modeling,tjahjowidodo2006nonlinear} (Fig.~\ref{fig:sea_bias_and_hysteresis}). Typically, this would require another system identification process. Instead however, an effort-based thermal model is introduced that automatically approximates the effects of bias and hysteresis on the actuator thermal dynamics as part of a single system identification process. This also provides the added benefit of not having to directly identify the nonlinear torque-current relationship of the actuators. Remembering that the torque output of a rotary joint powered by an electric motor is proportional to the product of the transmission ratio, $N$ and the torque-current constant, $K_m$, an equivalent thermal model of the same system instead uses the effort of the actuator as the input to the differential equation. Similarly, this linear relationship also holds true for an electric motor with a linear force output via a ball-screw mechanism. Generalizing the joint output effort, $F$, to represent torque for rotary outputs and forces for linear outputs, the relationship between actuator effort and electrical current can be described by
\begin{align}
\label{eq:effort_with_bias}
F = N K_m i_m + F_{\textrm{bias}}.
\end{align}

where $F_\textrm{bias}$ is nonzero for actuator thermal components with series-elastic actuators\footnote{A previously explored thermal model had ignored force bias and hysteresis effects. Surprisingly, this simpler model was also able to find parameters which can reasonably predict the evolution of the thermal states. While modeling the bias gives better fits and thermal prediction capabilities, the thermal recovery algorithm also works with the previous simpler model as it does not need exact thermal predictions.}. Notice that this current-effort relationship models the bias directly but only approximates the hysteresis with a linear fit. Next, note that the Joule heating of the electrical system is proportional to the square of the current, $P_e = i^2_m R_e$. Replacing this with an effort-based model by solving for $i_m$ in Eq.~\ref{eq:effort_with_bias}, the Joule heating based on the output joint effort is obtained:
\begin{align}
P_e &= (F^2 - 2F F_{\rm{bias}} + F^2_{\rm{bias}}) \frac{R_e}{(N K_m)^2} \\
    &= F^2 \beta - F\beta_{\rm{bias}} + P_{\rm{bias}} 
\end{align}
where $P_{\rm{bias}}$ is the constant Joule heating due to the actuator effort bias, and $\beta \triangleq R_e/(N K_m)^2$ and $\beta_{\rm{bias}} \triangleq 2F_{\rm{bias}} \beta$ are parameters that simultaneously encapsulate electrical resistance, transmission ratios, and actuator bias. Exploiting these constitutive relationships, Fig \ref{fig:current_vs_effort_thermal_models}(b) is an effort-based, first-order, thermal model of the motor driver and motor core systems. The dynamics of the effort-based thermal model are similar to the traditional current-based thermal model.
\begin{align}
\label{eq:effort_based_thermal_model}
RC \ \dot{T}(t) + T(t) = F^2 \beta R - F\beta_{\rm{bias}}R + T_{\text{offset}},
\end{align}
where $T_{\text{offset}} = T_{\rm{amb}} + P_{\rm{bias}}R$ is a temperature offset due to the ambient temperature and Joule heating bias. Note that the main advantage of using an effort-based thermal model is that the exact parameters for $N$, $K_m$, and $F_{\rm{bias}}$ are no longer needed for each actuator. Since this is a first order model, for an initial temperature $T_o$ and constant effort $F$, the state evolution has the following closed-form solution,
\begin{align}
T(t) = \ & T_o \ \textrm{e}^{\frac{-t}{RC}} + (F^2 \beta R - F\beta_{\rm{bias}}R + T_{\rm{offset}}) (1 - \textrm{e}^\frac{-t}{RC}).
\end{align}

\subsection{Thermal System Identification}
Operation data was gathered by placing Valkyrie in a variety of poses to get transient and steady-state thermal data on Valkyrie's legs and torso. A time-series data for temperature, electric current, and actuator effort for all actuators were obtained from this process. 
To identify the thermal parameters of the effort-based thermal model, a least-squares fit via batch gradient descent was used. Since at anytime, the system temperatures $T$, ambient temperature $T_{amb}$, and actuator effort $F$ are known, Eq.~\ref{eq:effort_based_thermal_model}, can be rearranged to express the unknown parameters. The $i$-th training data point $(x^i, y^i)$ can be written as 
\begin{align}
(T_i - T_{amb}) &= \begin{bmatrix} RC & \beta R & \beta_{\rm{bias}}R & P_{\rm{bias}}R \end{bmatrix} \begin{bmatrix} -\dot{T_i} \\ F_i^2 \\ -F_i \\ 1  \end{bmatrix}  \\
y^i &= \bm{\theta}^\top \bm{x}^i,
\end{align}
where $\theta = \begin{bmatrix}RC & \beta R & \beta_{\rm{bias}}R & P_{\rm{bias}}R \end{bmatrix}^\top$ is a vector of unknown parameters to be learned. Note that the training data, $x^i$, are low-pass filtered values of the raw data. Using standard regression techniques \cite{ngmachine}, the loss function, $L$, for this batch gradient descent and the corresponding update rule for the $j$-th parameter are,
\begin{align}
L &= \frac{1}{2m} \sum_{i=1}^m (\bm{\theta}^T \bm{x}^i - y^i)^2 \\
\frac{\partial L}{\partial \theta_j} &= \frac{1}{m} \sum_{i=1}^m (\bm{\theta}^T \bm{x}^i - y^i) x^i_{j} \\
\theta_j^{\rm{next}} &= \theta_j^{\rm{prev}} - \alpha \frac{\partial L}{\partial \theta_j},
\end{align}
where $\alpha$ is the learning rate, and $m$ is the number of training data. But, this naive formulation takes a long time to converge. To speed up the identification process, the Z-score is used for feature scaling except for the bias term, which ensures that the training data are transformed to variables with zero mean, \begin{align}
z^i_{(x,j)} = \frac{x^i_j - \overline{\bm{x}}_j}{\sigma_{(x,j)}}, \ z^i_y = \frac{y^i - \overline{\bm{y}}}{\sigma_y},
\end{align}
where, $(\overline{\cdot})$, is the mean of a vector valued variable and $\sigma_{(\cdot)}$ is its standard deviation. The $i$-th training data for the transformed problem is now written as
\begin{align}
z^i_y &=  \begin{bmatrix} \theta_{(z,1)} & \theta_{(z,2)} & \theta_{(z,3)} & P_{\rm{bias}}R \end{bmatrix}  \begin{bmatrix} z^i_{(x,1)} \\ z^i_{(x,2)} \\ z^i_{(x,3)} \\ 1 \end{bmatrix}, \label{eq:feature_scaled_model} \\
\bm{z}_y &= \bm{\theta}_z^T \bm{z}_x.
\end{align}
Batch gradient descent is then performed on this transformed data set with zero as the initial guess for all the parameters. Finally, solving for the untransformed variable $y$ in Eq.~\ref{eq:feature_scaled_model} reveals the relationship between the unscaled and scaled parameters as well as a learned temperature offset, $T_\textrm{offset}$, which models the Joule heating bias from the actuator and the true immediate ambient temperature of the thermal system from simply being powered on.
\begin{align}
\label{eq:scaled_equation_one}
y &= \sum_{j=1}^{3} \frac{\sigma_y}{\sigma_{(x,j)}} \theta_{(z,j)} x_j + T_\textrm{offset}, \\
T_\textrm{offset} &= (\overline{\bm{y}} -  \sum_{j=1}^{3} \frac{\sigma_y \overline{\bm{x}}_j}{\sigma_{(x,j)}} + P_{\rm{bias}}R). \nonumber
\end{align}
Namely, the relationship between the $j$-th unscaled parameter, $\theta_j$, and the scaled parameter, $\theta_{(z,j)}$, can be obtained with
\begin{align}
\theta_j = \frac{\sigma_y}{\sigma_{(x,j)}} \theta_{(z,j)}.
\label{eq:scaled_equation_two}
\end{align}
Combining Eq.~\ref{eq:scaled_equation_one} and Eq.~\ref{eq:scaled_equation_two} and substituting the original thermal variables, the identified thermal dynamics are
\begin{align}
T - T_{\textrm{amb}} &= -RC \ \dot{T} + F^2 \beta R - F \beta_{\rm{bias}}R + T_{\textrm{offset}}.
\label{eq:identified_thermal_dynamics}
\end{align}

\subsection{Thermal Prediction Performance}
\begin{figure*}
\centering
\includegraphics[width=2.\columnwidth]{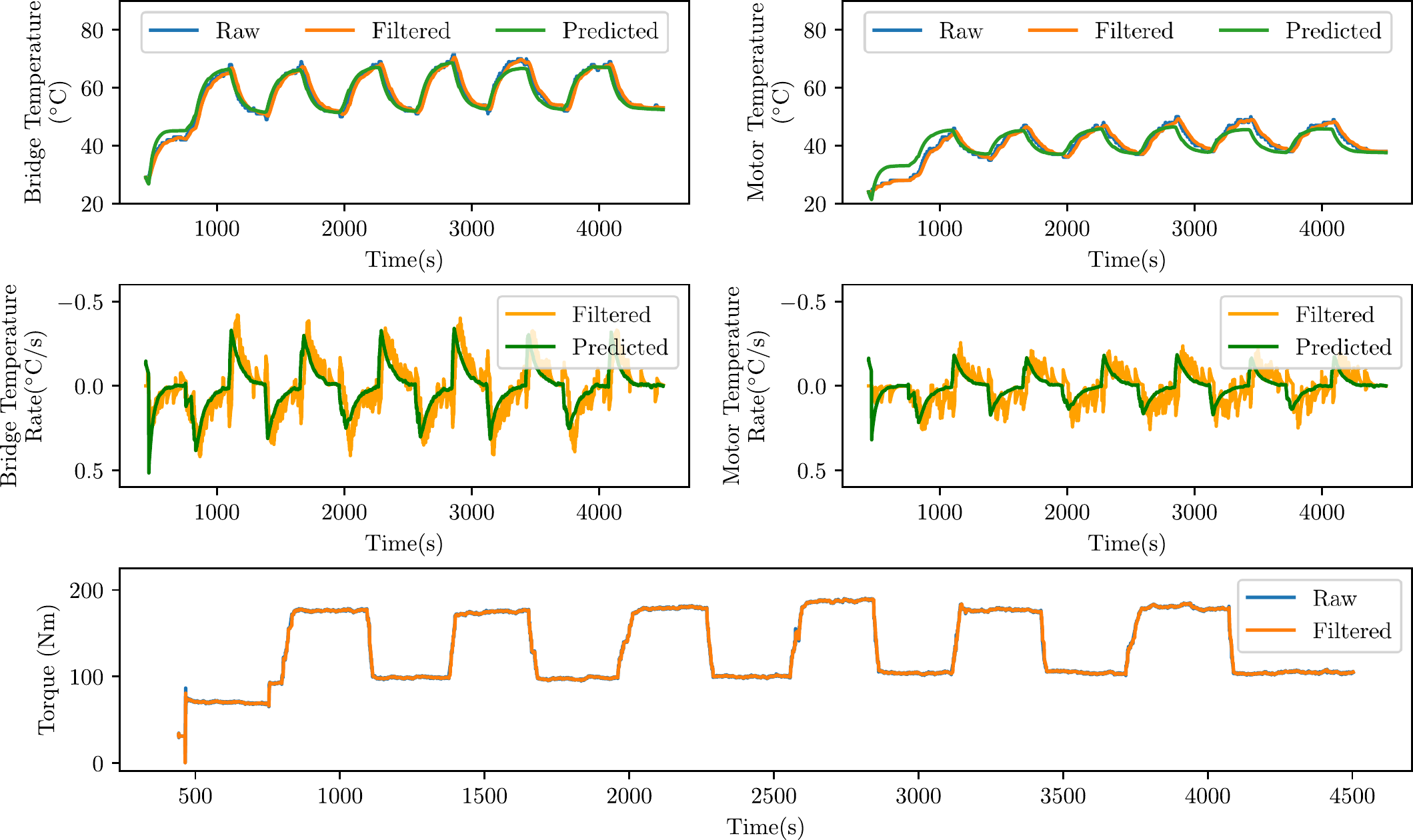}
\caption{A comparison between the true thermal states and predicted thermal states by the system identified model. The top graphs show the motor and bridge temperatures, motor and bridge temperature rates, and actuator effort data of Valkyrie's left knee actuator. The data was low-pass filtered with cutoff frequencies of 0.015Hz and 1Hz for temperature and torque measurements respectively. The temperature rates (middle graphs) were estimated using a low-pass derivative filter on the raw temperature measurement, also with a 0.015Hz cutoff frequency. Setting the initial condition of the thermal model to be equal to the ambient temperature (25$^\circ C$) and only using actuator effort as the input to the thermal model dynamics, the top plot shows that the predicted evolution of thermal states closely follows the true evolution of motor and bridge thermal states for the entire duration of the experiment without any sensed temperature updates on the thermal model.}
\label{fig:lj4_thermal_states}
\end{figure*}

The thermal model of an actuator is first initialized at ambient temperature, which was measured to be 25$^{\circ}C$. Using only the actuator effort as input to the internal thermal model, and without updating the measured temperature value, Euler integration is used to predict and simulate the evolution of the thermal states. The prediction performance is compared against operation data in which Valkyrie was repeatedly commanded to perform a double support squat and stand up in 4.5 minute intervals.
Fig.~\ref{fig:lj4_thermal_states} demonstrates that the thermal model and the learned parameters predict the thermal state evolution of the motor driver and motor core temperatures of Valkyrie's left knee actuator for very long time horizons without any sensed temperature updates.

\section{Finding Thermally Minimizing Configurations}
For a given a set of contact constraints, the thermal model is used to identify what contact-consistent configuration should be taken now, so that after some time, $\Delta t$, the overall system temperature is lowered. To find such a configuration, a fast, projection-based gradient-descent approach is presented.

For the following discussion, let $\bm{q} \in \mathbb{R}^n$ be the generalized coordinates of the floating-base robot with $n$ degrees of freedom, and $\bm{\Gamma} \in \mathbb{R}^m$ be the torque vector with $m < n$ actuated joints. A multi-limbed robot in contact with the environment has the following standard dynamics equations,
\begin{align}
\bm{A}\bm{\ddot{q}} + \bm{b} + \bm{g} = \bm{S}_a^\top\bm{\Gamma} + \bm{J}_c^\top \bm{F}_r,
\label{eq:robot_dynamics}
\end{align}
where $\bm{A}$, $\bm{b}$, and $\bm{g}$ are the inertia matrix, centrifugal forces, and gravitational forces respectively. $\bm{S}_a$ is a binary matrix which maps the torque vector to the corresponding generalized-coordinates. $\bm{J}_c$ is the contact Jacobian and $\bm{F}_r$ is the reaction force vector. A limb of the robot having a fixed contact constraint is defined by the following equation,
\begin{align}
\ddot{\bm{x}} = \bm{J}_c\ddot{\bm{q}} + \dot{\bm{J}}_c \dot{\bm{q}} = 0,
\label{eq:contact_constraint}
\end{align}
where $\bm{x}$ is the position and/or orientation of the contact, and $\bm{J}_c$ is its Jacobian. Substituting Eq.~\ref{eq:contact_constraint} to Eq.~\ref{eq:robot_dynamics}, and using the dynamically-consistent pseudo-inverse operator \cite{sentis2007synthesis}, $\overline{\bm{X}} = \bm{A}^{-1}\bm{X}^\top(\bm{X} \bm{A}^{-1} \bm{X}^\top)^{\dagger}$, for a matrix $\bm{X}$ with $^\dagger$ being the generalized pseudo-inverse, the following constrained dynamics equation is obtained.
\begin{align}
\bm{A}\bm{\ddot{q}} + \bm{N}_c^\top(\bm{b} + \bm{g}) + \bm{J}_c^\top(\bm{J}_c\bm{A}^{-1}\bm{J}_c^\top)^{-1}\bm{\dot{J}}_c\bm{\dot{q}}= (\bm{S}_a \bm{N}_c)^\top \bm{\Gamma} 
\label{eq:constrained_dynamics}
\end{align}
where $\bm{N}_c = (\bm{I} - \overline{\bm{J}}_c\bm{J}_c)$ is the contact null space. Notice that the contact reaction force has been eliminated from this equation, but can still be estimated with Eq.~\ref{eq:robot_dynamics} and the dynamically-consistent pseudo inverse operator.
\begin{align}
\bm{F}_r = \overline{\bm{J}_c^\top}(\bm{A}\bm{\ddot{q}} + \bm{b} + \bm{g} - \bm{S}_a^\top \bm{\Gamma}). \label{eq:contact_reaction_force}
\end{align}
Similarly, for any given dynamics on the left-hand side of Eq.~\ref{eq:constrained_dynamics}, a compensating, contact-consistent, actuator torque vector can be obtained,
\begin{align}
\bm{\Gamma} = (\overline{\bm{S}_a\bm{N_c}})^\top (\bm{A}\bm{\ddot{q}} + \bm{N}_c^\top(\bm{b} + \bm{g}) + \bm{J}_c^\top(\bm{J}_c\bm{A}^{-1}\bm{J}_c^\top)^{-1}\bm{\dot{J}}_c\bm{\dot{q}} ) .
\end{align}
Since the goal is to obtain a final minimizing configuration $(\dot{\bm{q}} = \bm{0})$, the compensating torque vector of interest is only dependent on configuration.
\begin{align}
\bm{\Gamma}(\bm{q}) = (\overline{\bm{S}_a\bm{N}_c(\bm{q})})^\top \bm{N}_c^\top(\bm{q}) \ \bm{g}(\bm{q}),
\label{eq:torque_from_gravity}
\end{align}
where the argument $\bm{q}$ is included for clarity. Finally, let $\bm{F}^\textrm{act}$ be the vector of the actuators' output efforts. A Jacobian, $\bm{J}_{\gamma}$, between the torque vector and the actuator effort vector always exists. Furthermore, Eq.~\ref{eq:torque_from_gravity} can be used to derive a direct relationship between robot configurations and actuator effort.
\begin{align}
\bm{F}^\textrm{act}(\bm{q}) &= \bm{J}_{\gamma}(\bm{q}) \bm{\Gamma}(\bm{q}).
\label{eq:actuator_effort_from_torques}
\end{align}
For rotary actuators with rotary outputs, the mapping between the actuator output effort and the joint torque is one-to-one. For joints with multiple actuators such as those found in Valkyrie's torso and ankle joints (See Sec.~\ref{sec:val_robot_system}), the mapping is dependent on the robot configuration $\bm{q}$. 
Since Eq.~\ref{eq:actuator_effort_from_torques} gives a relationship between the configuration of the robot and the actuator efforts needed to balance, this equation can be used to predict how the actuator temperatures after a time $\Delta t$ will change given a configuration $\bm{q}$. Let $\bm{T}(\bm{q}, \Delta t)$ be the temperature vector of actuators, with the $i$-th element being the $i$-th thermal dynamics with the closed-form solution of Eq.~\ref{eq:identified_thermal_dynamics},
\begin{eqnarray}
T_i & = T^i_o \ \textrm{e}^{\frac{-\Delta t}{(RC)_i}} \ + & \Big( F_j^2(\beta R)_i - F_j (\beta_{\rm{bias}})_i + T_{\rm{offset}} \Big) \cdot \nonumber \\  
& & \Big(1 - \textrm{e}^\frac{-\Delta t}{(RC)_i} \Big),
\end{eqnarray}
where $T^i_o$ is the initial temperature of the $i$-th system and $F_j(\bm{q})$ is the $j$-th actuator effort from $\bm{F}^\textrm{act}(\bm{q})$. This temperature vector provides the predicted actuator temperatures after a time $\Delta t$ for a given robot configuration. Since the goal is to find a thermally minimizing configuration, the following quadratic potential function is constructed,
\begin{align}
f(\bm{q}) = \bm{T}(\Delta t, \bm{q})^\top \cdot \bm{Q} \cdot \bm{T}(\Delta t, \bm{q}), 
\label{eq:temperature_potential_function}
\end{align}
where $\bm{Q}$ is a diagonal cost matrix which can be modified online after receiving a temperature update to prioritize the minimization of some thermal states over the others.

Next, a minimizing configuration can be obtained by iteratively computing the contact-consistent gradient of Eq.~\ref{eq:temperature_potential_function} with
\begin{align}
\nabla f(\bm{q}) & = \begin{bmatrix} \frac{\partial f(\bm{q})}{\partial q_1}, & \frac{\partial f(\bm{q})}{\partial q_2}, ..., \frac{\partial f(\bm{q})}{\partial q_n} \end{bmatrix}, \label{eq:potential_gradient_main}
\\
 \frac{\partial f(\bm{q})}{\partial q_i} &\approx \frac{f(N_c(\bm{q} + h\cdot\bm{\epsilon}_i)) - f(\bm{q})}{h}, \textrm{ for } i = 1, ..., n,
\label{eq:potential_gradient}
\end{align}
for a small value of $h$ with $\bm{\epsilon}_i$ being a zero vector with a $1$ at the $i$-th element. A way to interpret Eq.~\ref{eq:potential_gradient} is the following. Since $\bm{N}_c$ is a projector matrix $(\bm{N}^2_c= \bm{N}_c)$\cite{sentis2007synthesis}, for a given configuration proposal, $\bm{q}+h\cdot\bm{\epsilon}$, the proposed new configuration is projected with $\bm{N}_c$ to ensure kinematic contact constraint satisfiability before evaluating the temperature potential, $f(\bm{q})$. Another approach to constructing this gradient is to first find the set of basis vectors of the null space, $V = \{ \bm{v}_1, \bm{v}_2, ... | \bm{N}_c \bm{v}_i = 0 \}$, which already satisfy the kinematic contact constraints, then find the directional gradient,
\begin{align}
\nabla f(\bm{q}) & = \sum_{i=1}^{|V|} \bm{v}_i \cdot \frac{\partial f(\bm{q})}{\partial \bm{v}_i},  \\
 \frac{\partial f(\bm{q})}{\partial \bm{v}_i} &\approx \frac{f(\bm{q} + h\cdot\bm{v}_i) - f(\bm{q})}{h}, \textrm{ for } i = 1, ..., |V|.
\end{align}
This second approach has the advantage of having a smaller number of basis vectors as the number of contact constraints increases. In either case, the contact-consistent gradient can be used to iteratively update the configuration $\bm{q}$ with
\begin{align}
\bm{dq} &= -\bm{k}_p \nabla f(\bm{q}) \label{eq:dq_update} \\
\bm{q}_{k+1} &= \bm{q}_k + \bm{N}_c \bm{dq}, \label{eq:configuration_update}
\end{align}
where $\bm{k}_p$ is a vector descent gain. In this implementation, the gain changes with the magnitude of $\nabla f(\bm{q})$. At every iteration, values of $\bm{k}_p$ are selected such that the maximum configuration change is bounded by some $\delta_{\textrm{max}}$. For instance, looking at the maximum actuated joint configuration change, $\textrm{max}(\bm{dq}_\textrm{act})$, the actuated joint gains are selected to be $\bm{k}^{\textrm{act}}_p = \delta_{\rm{max}}/\textrm{max}(\bm{dq}_\textrm{act}) $. The same gain scaling method is also performed for the linear and rotary components of the floating base configurations. While the potential function is convex, the descent algorithm finds solutions that are close to, but not the true minimum, as the naive implementation uses a fixed $\delta_{\textrm{max}}$ step size and terminates when the iteration limit is hit or if the next iterate causes the cost to increase.

\begin{figure*}
\includegraphics[width=2.\columnwidth]{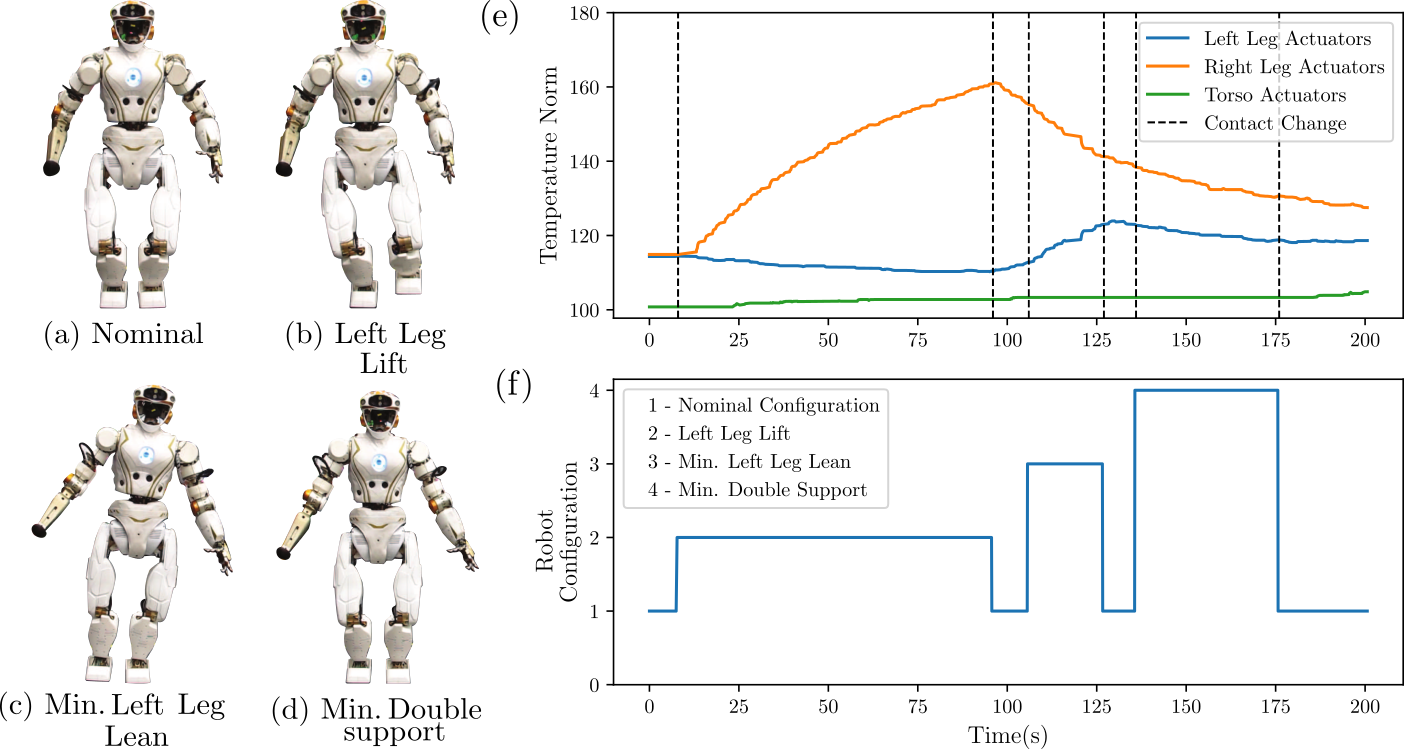}
\caption{Experimental validation of the actuator thermal recovery algorithm. Subfigures(a)-(d) show the four different configurations Valkyrie entered during this experiment. Configurations (a) and (b) are prescribed while the minimizing configurations, (c) and (d), were found by the thermal minimization algorithm. Subfigure (e) shows the temperature $l^2$-norms of the left leg, right leg, and torso actuators. Subfigure (f) shows the configuration of Valkyrie as a function of time. In this experiment, Valkyrie was forced to balance on its right leg causing the right leg actuators to heat up. When at least one of the thermal states of the right leg actuators exceeded $75^\circ C$ , the thermal minimization algorithm is called. The algorithm first finds a minimizing configuration with only a left foot contact. While this heats up the left leg actuators, this also significantly cools down the right leg actuators. After 20 seconds, it finds a new minimizing configuration and enters a double support contact configuration, which simultaneously cools down the left and right leg actuators. Finally, Valkyrie is returned to a nominal configuration once all of the thermal states have safely been recovered from unsafe regions. All configuration changes go through the nominal configuration as part of the assumption that a valid trajectory exists between contact modes.}
\label{fig:left_leg_experiment}
\end{figure*}

\section{Actuator Thermal Recovery Algorithm}
Given the above derivations, an actuator thermal recovery algorithm for a multi-limbed robot having a number of valid contact configurations can now be constructed. Let the set $C$ be a collection of contact configurations, $c$. For instance, the set $C$ for Valkyrie standing in place, would contain three elements: $c_1 = \textrm{double support}$, $c_2 = \textrm{left leg support}$, and $c_3 = \textrm{right leg support}$. At every control interval, the strategy to thermally recover the robot's actuators solves the following minimization problem, 
\begin{align}
(c,\bm{q})=
\underset{c \in C }{\textrm{argmin}} \big( \underset{\bm{q}}{\text{argmin}} f(\bm{q}, c) \big),
\label{eq:min_algorithm}
\end{align}
where $f(\bm{q}, c)$ is Eq.~\ref{eq:temperature_potential_function} but with the contact configuration included for clarity.
In other words, from a set of contact configurations $C$, select which contact configuration, $c$, and corresponding robot configuration $\bm{q}$ best minimizes the temperature potential function. This enables a contact-switching strategy with temperature minimization, which is the key to cooling dangerously warm actuators faster than a strategy relying on minimum effort configuration only. Finally, it is assumed that there exists a valid trajectory to switch between any contact configurations in $C$. While Eq.~\ref{eq:temperature_potential_function} takes about 15s per configuration to converge due to the naive gradient descent implementation, since the contact configurations are simple, the final minimizing configurations turn out to be similar and can be stored ahead of time.

To encourage the optimization routine to select contact configurations which prioritize thermal minimization of dangerously warm actuator components, The $i$-th diagonal element of $\bm{Q}$ is updated to a large number if a sensed thermal state is greater than some threshold (eg: 70$^\circ C$). Otherwise, it is set to its original weight. Additionally, the thermal prediction horizon is set to half of the learned thermal time constant, $RC$. This ensures that the predicted thermal state is still in the transient region and not in steady-state. Note that setting the prediction horizon to infinity is similar to finding a minimizing configuration with as many active contacts as possible, which is not desired as contact switching can cool down very warm actuators faster (see Sec.~\ref{sec:approach_norm_rate_comparisons}). The implementation uses $\Delta t = 20\rm{s},$ and the $Q$ cost matrix is identity by default.

\section{Experimental Validation and Results}
\subsection{Thermal Recovery with Contact Switching}
The NASA Valkyrie robot is used to validate the thermal minimization algorithm. The robot uses the Institute for Human and Machine Cognition's (IHMC) momentum-based whole-body controller \cite{koolen2016design} with a high level interface\footnote{\url{https://github.com/ihmcrobotics/ihmc_msgs}}. To begin the experiment, the operator first commands Valkyrie to balance on a single leg. This heats up the stance leg in less than two minutes. When the thermal state of the torso or leg actuators enters a warning zone (set to 75$^\circ C$), the thermal recovery algorithm routine is called\footnote{Valkyrie's actuators can operate up to 90$^\circ C$ but a low value is set here for hardware safety.}. 

In Fig.~\ref{fig:left_leg_experiment}, the left leg is lifted, which heats up the right leg actuators. To thermally recover the heated right leg actuators, the thermal minimization algorithm finds a strategy to first balance on the left leg, then balance with both legs after the right leg actuators have sufficiently cooled down. Notice that the descent of the temperature norm of the right leg actuators was steeper during left leg balancing than double support balancing, indicating that the algorithm correctly selected  a strategy that managed the heating and cooling of the left and right leg actuators respectively to prioritize the cooling of the right leg actuators. When all the thermal states are below 70$^\circ C$, the robot is returned to its nominal configuration.

\subsection{Proposed Strategy vs Minimum Effort Strategy}
\label{sec:approach_norm_rate_comparisons}
\begin{figure}
\includegraphics[width=\columnwidth]{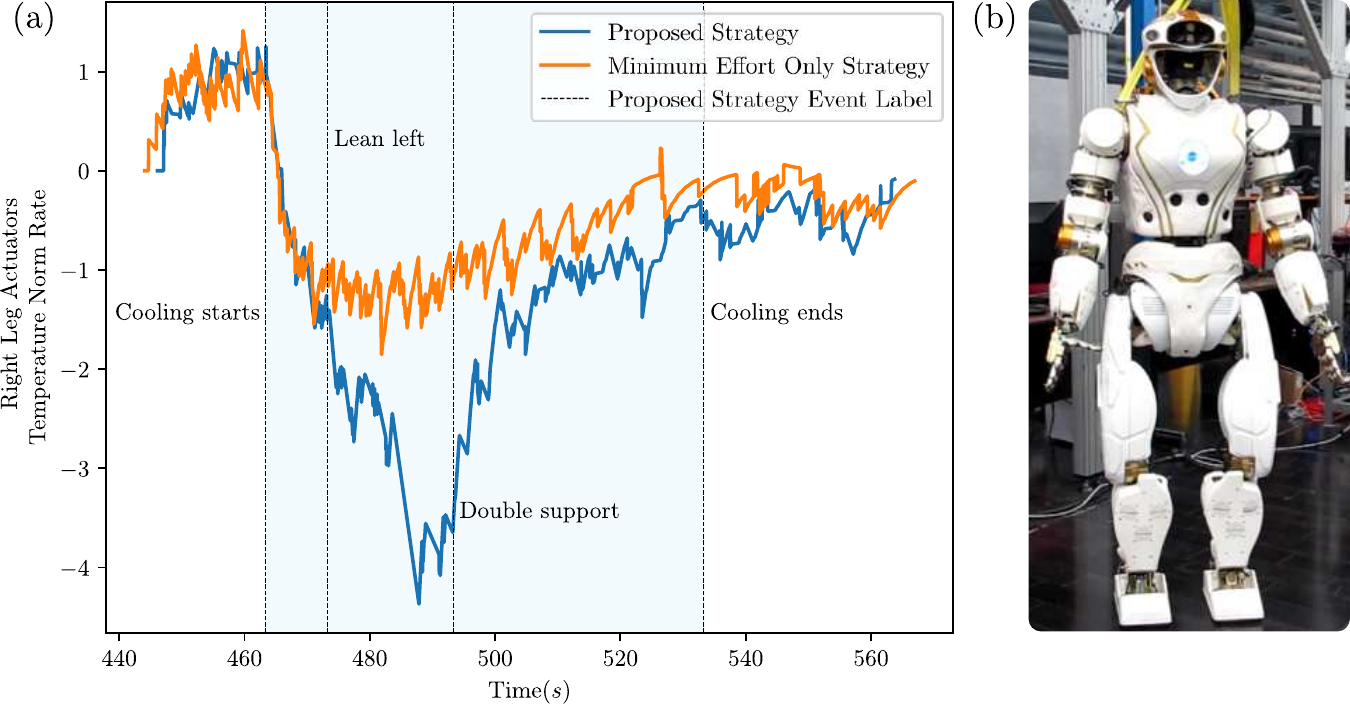}
\caption{(a) The temperature norm rates of the dangerously warm right leg actuators are compared between the proposed thermally-aware contact switching approach versus minimum effort only. The norm rates are obtained by performing a low-pass derivative filter on the temperature norm with a 0.1Hz cutoff frequency. The data are x-axis aligned by using the starting time of the thermal recovery process. (b) The minimum effort configuration used for comparison. While both approaches finished cooling all the actuators in one minute, the contact-switching strategy had faster cooling rates for the right leg actuator, cooling it down to a safe temperature state in 21s (which is the duration between when the robot leans on its left leg and when it enters double support).} 
\label{fig:approach_norm_rate_comparison}
\end{figure}

The advantage of using a thermally-aware contact-switching strategy instead of a minimum effort strategy is highlighted by temperature norm changes for the right leg actuators (Fig.\ref{fig:approach_norm_rate_comparison}).  The previous experiment is repeated but when an actuator's thermal state enters the warning zone, the robot is instead commanded to only employ a minimum effort configuration. This configuration is computed using Eq.~\ref{eq:torque_from_gravity} as the vector term in Eq.~\ref{eq:temperature_potential_function}. As before, the robot is returned to a nominal configuration once all the actuators are below $70^\circ C$. Fig.\ref{fig:approach_norm_rate_comparison} shows that while both approaches take about one minute to recover the actutator thermal states, the contact-switching approach has faster cooling rates for the dangerously warm right leg actuators. This enables the right leg actuators to cool down in only 21s (the duration between when the robot leans on the left leg and when the robot enters double support).

\section{Discussions and Conclusions}
The presented approach assumes that the estimated contact reaction forces (Eq.~\ref{eq:contact_reaction_force}) are always valid as gradient descent is performed on the configuration. Since, surface contacts typically have unilateral constraints, the gradient descent may propose a configuration that requires an invalid reaction force direction. To address this problem, observe that Eq.~\ref{eq:dq_update} is equivalent to a joint configuration task. Thus, this can be inserted in a QP-based whole-body controller with task relaxation, such as \cite{kim2018computationally}, which ensures unilateral constraint satisfiability. This QP can then be a subroutine to compute the configuration update (Eq.~\ref{eq:configuration_update}) with numerical integration of the inverse dynamics, as used in \cite{kim2016dynamic} for example.

The approach for thermally recovering actuators uses data-driven methods and closed-form solutions of equality constrained dynamics. While the thermally minimizing configurations found usually coincide with minimum effort, significantly different thermal states and time constants along the same kinematic chain can generate a different solution. Furthermore, contact switching strategies enable the robot to aggressively cool down dangerously warm actuators faster than a minimum effort strategy alone. A future research direction is to incorporate the closed-form thermal prediction as part of general trajectory generation and motion planning.

\section*{Acknowledgments}
This work was supported by a NASA Space Technology Research Fellowship (NSTRF) Grant \#NNX15AQ42H and by the Office of Naval Research, ONR grant \#N000141512507. The authors are grateful to the Valkyrie team at NASA Johnson Space Center for providing support on robot maintenance and operation, the Institute for Human and Machine Cognition (IHMC) for their open-sourced control algorithm, and the members of the Human-Centered Robotics Lab (HCRL) at UT Austin for their support and insights.

\ifCLASSOPTIONcaptionsoff
  \newpage
\fi



%

\bibliographystyle{IEEEtran}

\begin{thebibliography}{10}
\providecommand{\url}[1]{#1}
\csname url@rmstyle\endcsname
\providecommand{\newblock}{\relax}
\providecommand{\bibinfo}[2]{#2}
\providecommand\BIBentrySTDinterwordspacing{\spaceskip=0pt\relax}
\providecommand\BIBentryALTinterwordstretchfactor{4}
\providecommand\BIBentryALTinterwordspacing{\spaceskip=\fontdimen2\font plus
\BIBentryALTinterwordstretchfactor\fontdimen3\font minus
  \fontdimen4\font\relax}
\providecommand\BIBforeignlanguage[2]{{%
\expandafter\ifx\csname l@#1\endcsname\relax
\typeout{** WARNING: IEEEtran.bst: No hyphenation pattern has been}%
\typeout{** loaded for the language `#1'. Using the pattern for}%
\typeout{** the default language instead.}%
\else
\language=\csname l@#1\endcsname
\fi
#2}}

\bibitem{han2016protection}
Y.~Han, W.~Luan, Y.~Jiang, and X.~Zhang, ``Protection of electronic devices on
  nuclear rescue robot: Passive thermal control,'' \emph{Applied Thermal
  Engineering}, vol. 101, pp. 224--230, 2016.

\bibitem{swanson2003nasa}
T.~D. Swanson and G.~C. Birur, ``Nasa thermal control technologies for robotic
  spacecraft,'' \emph{Applied thermal engineering}, vol.~23, no.~9, pp.
  1055--1065, 2003.

\bibitem{radford2015valkyrie}
N.~A. Radford, P.~Strawser, K.~Hambuchen, J.~S. Mehling, W.~K. Verdeyen, A.~S.
  Donnan, J.~Holley, J.~Sanchez, V.~Nguyen, L.~Bridgwater, \emph{et~al.},
  ``Valkyrie: Nasa's first bipedal humanoid robot,'' \emph{Journal of Field
  Robotics}, vol.~32, no.~3, pp. 397--419, 2015.

\bibitem{seok2015design}
S.~Seok, A.~Wang, M.~Y.~M. Chuah, D.~J. Hyun, J.~Lee, D.~M. Otten, J.~H. Lang,
  and S.~Kim, ``Design principles for energy-efficient legged locomotion and
  implementation on the mit cheetah robot,'' \emph{IEEE/ASME Transactions on
  Mechatronics}, vol.~20, no.~3, pp. 1117--1129, 2015.

\bibitem{kenneally2016design}
G.~D. Kenneally, A.~De, and D.~E. Koditschek, ``Design principles for a family
  of direct-drive legged robots.'' \emph{IEEE Robotics and Automation Letters},
  vol.~1, no.~2, pp. 900--907, 2016.

\bibitem{urata2008thermal}
J.~Urata, T.~Hirose, Y.~Namiki, Y.~Nakanishi, I.~Mizuuchi, and M.~Inaba,
  ``Thermal control of electrical motors for high-power humanoid robots,'' in
  \emph{Intelligent Robots and Systems, 2008. IROS 2008. IEEE/RSJ International
  Conference on}.\hskip 1em plus 0.5em minus 0.4em\relax IEEE, 2008, pp.
  2047--2052.

\bibitem{paine2015design}
N.~Paine and L.~Sentis, ``Design and comparative analysis of a retrofitted
  liquid cooling system for high-power actuators,'' in \emph{Actuators},
  vol.~4, no.~3.\hskip 1em plus 0.5em minus 0.4em\relax Multidisciplinary
  Digital Publishing Institute, 2015, pp. 182--202.

\bibitem{englsberger2014overview}
J.~Englsberger, A.~Werner, C.~Ott, B.~Henze, M.~A. Roa, G.~Garofalo, R.~Burger,
  A.~Beyer, O.~Eiberger, K.~Schmid, \emph{et~al.}, ``Overview of the
  torque-controlled humanoid robot toro,'' in \emph{Humanoid Robots
  (Humanoids), 2014 14th IEEE-RAS International Conference on}.\hskip 1em plus
  0.5em minus 0.4em\relax IEEE, 2014, pp. 916--923.

\bibitem{dafarra2016torque}
S.~Dafarra, F.~Romano, and F.~Nori, ``Torque-controlled stepping-strategy push
  recovery: Design and implementation on the icub humanoid robot,'' in
  \emph{Humanoid Robots (Humanoids), 2016 IEEE-RAS 16th International
  Conference on}.\hskip 1em plus 0.5em minus 0.4em\relax IEEE, 2016, pp.
  152--157.

\bibitem{krotkov2017darpa}
E.~Krotkov, D.~Hackett, L.~Jackel, M.~Perschbacher, J.~Pippine, J.~Strauss,
  G.~Pratt, and C.~Orlowski, ``The darpa robotics challenge finals: results and
  perspectives,'' \emph{Journal of Field Robotics}, vol.~34, no.~2, pp.
  229--240, 2017.

\bibitem{guizzo2015hard}
E.~Guizzo and E.~Ackerman, ``The hard lessons of darpa's robotics challenge
  [news],'' \emph{IEEE Spectrum}, vol.~52, no.~8, pp. 11--13, 2015.

\bibitem{sentis2007synthesis}
L.~Sentis, ``Synthesis and control of whole-body behaviors in humanoid
  systems,'' Ph.D. dissertation, Stanford University, 2007.

\bibitem{stemme1994principles}
O.~Stemme and P.~Wolf, ``Principles and properties of highly dynamic dc
  miniature motors,'' \emph{Maxon Motor: Sachseln, Switzerland}, 1994.

\bibitem{duchi2011adaptive}
J.~Duchi, E.~Hazan, and Y.~Singer, ``Adaptive subgradient methods for online
  learning and stochastic optimization,'' \emph{Journal of Machine Learning
  Research}, vol.~12, no. Jul, pp. 2121--2159, 2011.

\bibitem{aghili2016control}
F.~Aghili and C.-Y. Su, ``Control of constrained robots subject to unilateral
  contacts and friction cone constraints,'' in \emph{Robotics and Automation
  (ICRA), 2016 IEEE International Conference on}.\hskip 1em plus 0.5em minus
  0.4em\relax IEEE, 2016, pp. 2347--2352.

\bibitem{pratt1995series}
G.~A. Pratt and M.~M. Williamson, ``Series elastic actuators,'' in
  \emph{Intelligent Robots and Systems 95.'Human Robot Interaction and
  Cooperative Robots', Proceedings. 1995 IEEE/RSJ International Conference on},
  vol.~1.\hskip 1em plus 0.5em minus 0.4em\relax IEEE, 1995, pp. 399--406.

\bibitem{paine2014design}
N.~Paine, S.~Oh, and L.~Sentis, ``Design and control considerations for
  high-performance series elastic actuators,'' \emph{IEEE/ASME Transactions on
  Mechatronics}, vol.~19, no.~3, pp. 1080--1091, 2014.

\bibitem{ruderman2012modeling}
M.~Ruderman, ``Modeling of elastic robot joints with nonlinear damping and
  hysteresis,'' in \emph{Robotic Systems-Applications, Control and
  Programming}.\hskip 1em plus 0.5em minus 0.4em\relax InTech, 2012.

\bibitem{tjahjowidodo2006nonlinear}
T.~Tjahjowidodo, F.~Al-Bender, H.~Van~Brussel, \emph{et~al.}, ``Nonlinear
  modelling and identification of torsional behaviour in harmonic drives,'' in
  \emph{Proc. International Conference on Noise and Vibration Engineering
  (ISMA2006)}.\hskip 1em plus 0.5em minus 0.4em\relax Citeseer, 2006, pp.
  2785--2796.

\bibitem{ngmachine}
A.~Ng, ``Machine learning,'' \emph{www.coursera.org/learn/machine-learning/.
  Produced at Standford University for Coursera}, 2014.

\bibitem{koolen2016design}
T.~Koolen, S.~Bertrand, G.~Thomas, T.~De~Boer, T.~Wu, J.~Smith, J.~Englsberger,
  and J.~Pratt, ``Design of a momentum-based control framework and application
  to the humanoid robot atlas,'' \emph{International Journal of Humanoid
  Robotics}, vol.~13, no.~01, p. 1650007, 2016.

\bibitem{kim2018computationally}
D.~Kim, J.~Lee, O.~Campbell, H.~Hwang, and L.~Sentis, ``Computationally-robust
  and efficient prioritized whole-body controller with contact constraints,''
  \emph{arXiv preprint arXiv:1807.01222}, 2018.

\bibitem{kim2016dynamic}
D.~Kim, S.~J. Jorgensen, P.~Stone, and L.~Sentis, ``Dynamic behaviors on the
  nao robot with closed-loop whole body operational space control,'' in
  \emph{Humanoid Robots (Humanoids), 2016 IEEE-RAS 16th International
  Conference on}.\hskip 1em plus 0.5em minus 0.4em\relax IEEE, 2016, pp.
  1121--1128.

\end{thebibliography}


\end{document}